\documentclass[letterpaper]{article} % DO NOT CHANGE THIS
\usepackage{aaai24}  % DO NOT CHANGE THIS
\usepackage{times}  % DO NOT CHANGE THIS
\usepackage{helvet}  % DO NOT CHANGE THIS
\usepackage{courier}  % DO NOT CHANGE THIS
\usepackage[hyphens]{url}  % DO NOT CHANGE THIS
\usepackage{graphicx} % DO NOT CHANGE THIS
\usepackage{natbib}  % DO NOT CHANGE THIS AND DO NOT ADD ANY OPTIONS TO IT
\usepackage{caption} % DO NOT CHANGE THIS AND DO NOT ADD ANY OPTIONS TO IT
\usepackage{booktabs}
\usepackage{amssymb}
\usepackage{amsmath}
\usepackage{multicol}
\usepackage{multirow}
\usepackage{subfigure}
\usepackage{algorithm}
\usepackage{algorithmic}

\usepackage{newfloat}
\usepackage{listings}
\DeclareCaptionStyle{ruled}{labelfont=normalfont,labelsep=colon,strut=off} % DO NOT CHANGE THIS
\floatstyle{ruled}
\newfloat{listing}{tb}{lst}{}
\floatname{listing}{Listing}
\title{Debiased Novel Category Discovering and Localization}
\author {
    Juexiao Feng\textsuperscript{\rm 1,2,3}\equalcontrib,
    Yuhong Yang\textsuperscript{\rm 1,2,3}\equalcontrib,
    Yanchun Xie\textsuperscript{\rm 4},
    Yaqian Li\textsuperscript{\rm 4},
    Yandong Guo\textsuperscript{\rm 4},
    Yuchen Guo\textsuperscript{\rm 1,2},
    Yuwei He\textsuperscript{\rm 1,2},
    Liuyu Xiang\textsuperscript{\rm 5}\thanks{Corresponding Authors.},
    Guiguang Ding\textsuperscript{\rm 1,2}\footnotemark[2]
}
\affiliations {
    \textsuperscript{\rm 1}Tsinghua University \\
    \textsuperscript{\rm 2}BNRist\\
    \textsuperscript{\rm 3}Hangzhou Zhuoxi Institute of Brain and Intelligence \\
    \textsuperscript{\rm 4}OPPO Research Institute\\
    \textsuperscript{\rm 5}Beijing University of Posts and Telecommunications\\
    \{fjx20, yangyh22\}@mails.tsinghua.edu.cn,
    \{xieyanchun, liyaqian\}@oppo.com,
    yandong.guo@live.com, \\
    \{yuchen.w.guo, heyuwei403\}@gmail.com,
    xiangly@bupt.edu.cn,
    dinggg@tsinghua.edu.cn
}

\begin{document}

\maketitle

\begin{abstract}
In recent years, object detection in deep learning has experienced rapid development. However, most existing object detection models perform well only on closed-set datasets, ignoring a large number of potential objects whose categories are not defined in the training set. These objects are often identified as background or incorrectly classified as pre-defined categories by the detectors. In this paper, we focus on the challenging problem of Novel Class Discovery and Localization (NCDL), aiming to train detectors that can detect the categories present in the training data, while also actively discover, localize, and cluster new categories. We analyze existing NCDL methods and identify the core issue: object detectors tend to be biased towards seen objects, and this leads to the neglect of unseen targets. To address this issue, we first propose an Debiased Region Mining (DRM) approach that combines class-agnostic Region Proposal Network (RPN) and class-aware RPN in a complementary manner. Additionally, we suggest to improve the representation network through semi-supervised contrastive learning by leveraging unlabeled data. Finally, we adopt a simple and efficient mini-batch K-means clustering method for novel class discovery. We conduct extensive experiments on the NCDL benchmark, and the results demonstrate that the proposed DRM approach significantly outperforms previous methods, establishing a new state-of-the-art.
\end{abstract}

\section{Introduction}

Object detection~\cite{DBLP:journals/corr/GirshickDDM13,DBLP:journals/corr/Girshick15,DBLP:journals/corr/RedmonDGF15} technology is a fundamental research area of computer vision and is now widely used in security~\cite{he2022secret}, autonomous driving~\cite{DBLP:journals/corr/abs-2104-13921}, medical~\cite{guo2022deep} field and other areas.

\begin{figure}[t]
  \centering
   \includegraphics[width=1\linewidth]{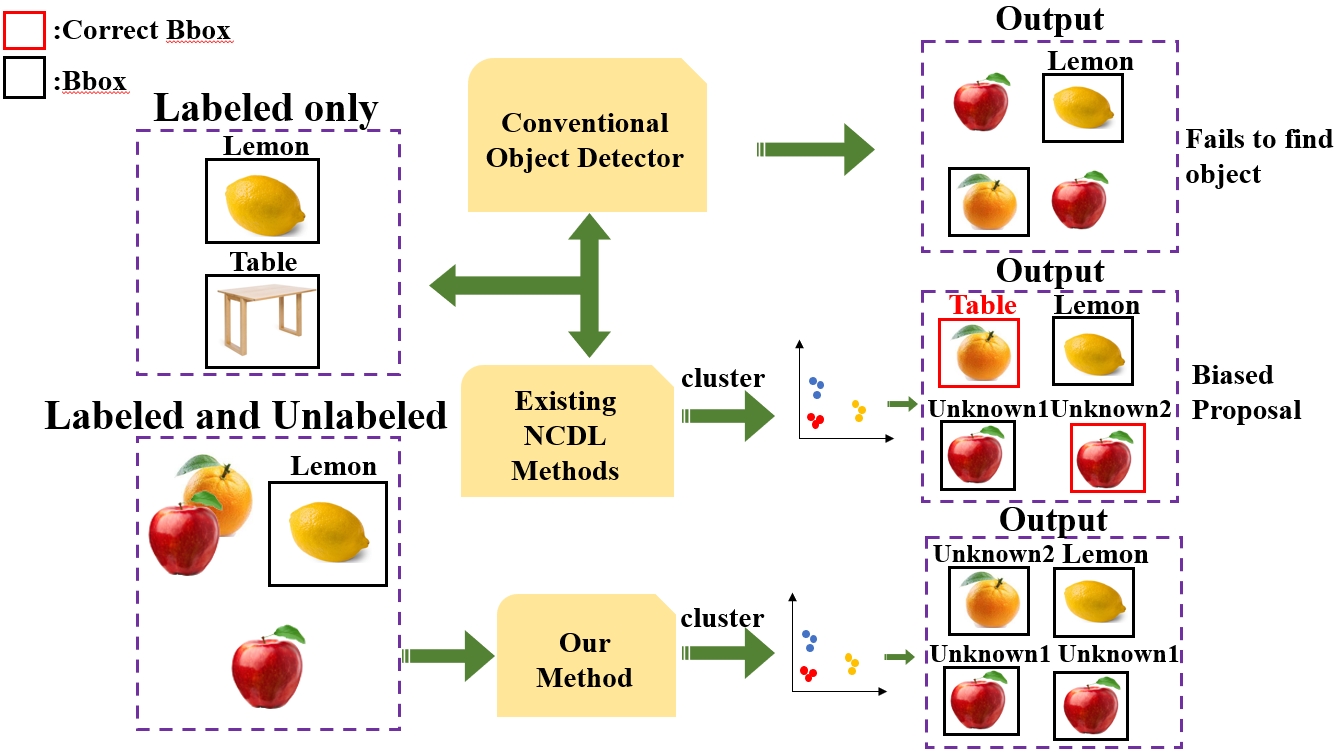}
   \caption{For current mainstream object detection models, they do not detect new objects in unlabeled data. The existing NCDL methods tend to extract biased proposals for unlabeled data. Our proposed model not only localizes unknown objects but also clusters different unknown objects, thereby achieving the discovery of new categories.}
   \label{fig:onecol}
\end{figure}

Existing object detection methods are trained on closed-set datasets with a fixed number of categories. The majority of research focuses on improving detection performance on closed-set datasets.
In closed-set datasets scenarios, object detectors are evaluated on a test set that contains the same set of categories as the training set. However, in the in real-world scenario, object detectors are faced with both known objects and potential unknown objects that have never appeared or been annotated in the training set. 
Once trained, they are unaware of any objects that they have not seen during training. They either treat the unknown objects as background or misclassify them as known categories. 
By contrast, humans have the capability of perceiving, discovering, and recognizing unknown novel objects. Therefore, the Novel Category Discovery problem, which aims to not only detect known objects, but also discover novel categories unsupervisedly, has attracted much attention

Most NCD methods involve a pre-training step on labeled sets and then perform on unlabeled data. For example,~\cite{han2019learning} adds partial information from known classes to unknown classes, pretending that some parts are unlabeled, and then runs clustering algorithms in the expansion phase for model inference. This approach of handling unlabeled data during pre-training will cause the model to learn from unknown data. While this can improve the model's ability to discover unknown data, the model should not have prior knowledge of the unlabeled dataset, which is not applicable in real-world scenarios. 
\cite{han2020automatically} utilizes all data, including labeled and unlabeled data, for pre-training. However, such methods require repetitive pre-training on all data after modifying the unlabeled set, resulting in expensive training costs.
Most NCD methods~\cite{han2021autonovel,han2019learning} have three steps. First, the object detector is pre-trained on a closed set. Second, localize and discover potential unknown new objects. Finally, the discovered objects are categorized into new categories by clustering. While this approach proves its effectiveness, most methods only utilize known objects and categories for pre-training and localization, which may introduce bias. Such methods use detection heads for closed sets which may introduce biased feature representations. Similarly, training RPNs only on labeled closed-set objects may also lead to biased localization.

To address the aforementioned issues, we propose debiased NCD methods to mitigate bias in both feature representation and object localization. Specifically, from the localization perspective, we introduce a self-supervised contrastive learning method. This method enables the model to learn similar features for similar data instances and differentiate unknown class objects from known class objects. Regarding feature representation, we propose dual RPN strategy to simultaneously detect target objects in the image. One RPN is designed to be class-aware, which aims to obtain accurate localization information for known classes. The other RPN is designed to be class-agnostic, where the classification loss is replaced with IOU regression to locate unlabeled target objects. The resulting bounding boxes from both detectors are selected as reliable confidence intervals, leading to improved performance for both known and unknown classes.

Finally, we conduct extensive experiments on NCDL benchmarks and demonstrate that the proposed debiased region mining method outperforms previous methods significantly, establishing a new state-of-the-art. 

Our contribution can be summarized as follows:
\begin{itemize}
    \item[$\bullet$]We revisit the problem of novel category discovery in an open world and investigate the bias problem in existing methods.
    \item[$\bullet$]We propose using dual object detectors together to obtain a good area proposal that can effectively find all target objects in the image and locate them better. 
    \item[$\bullet$]We design a Semi-Supervised Instance-level Contrastive Learning method to obtain image representations better than before and make the model rely on unlabeled image information to learn image features. 
    \item[$\bullet$]Based on this approach we have conducted a large number of experiments, which show that our method outperforms other baseline methods.
\end{itemize}

\begin{figure*}[t]
    \centering
    \includegraphics[width=0.8\linewidth]{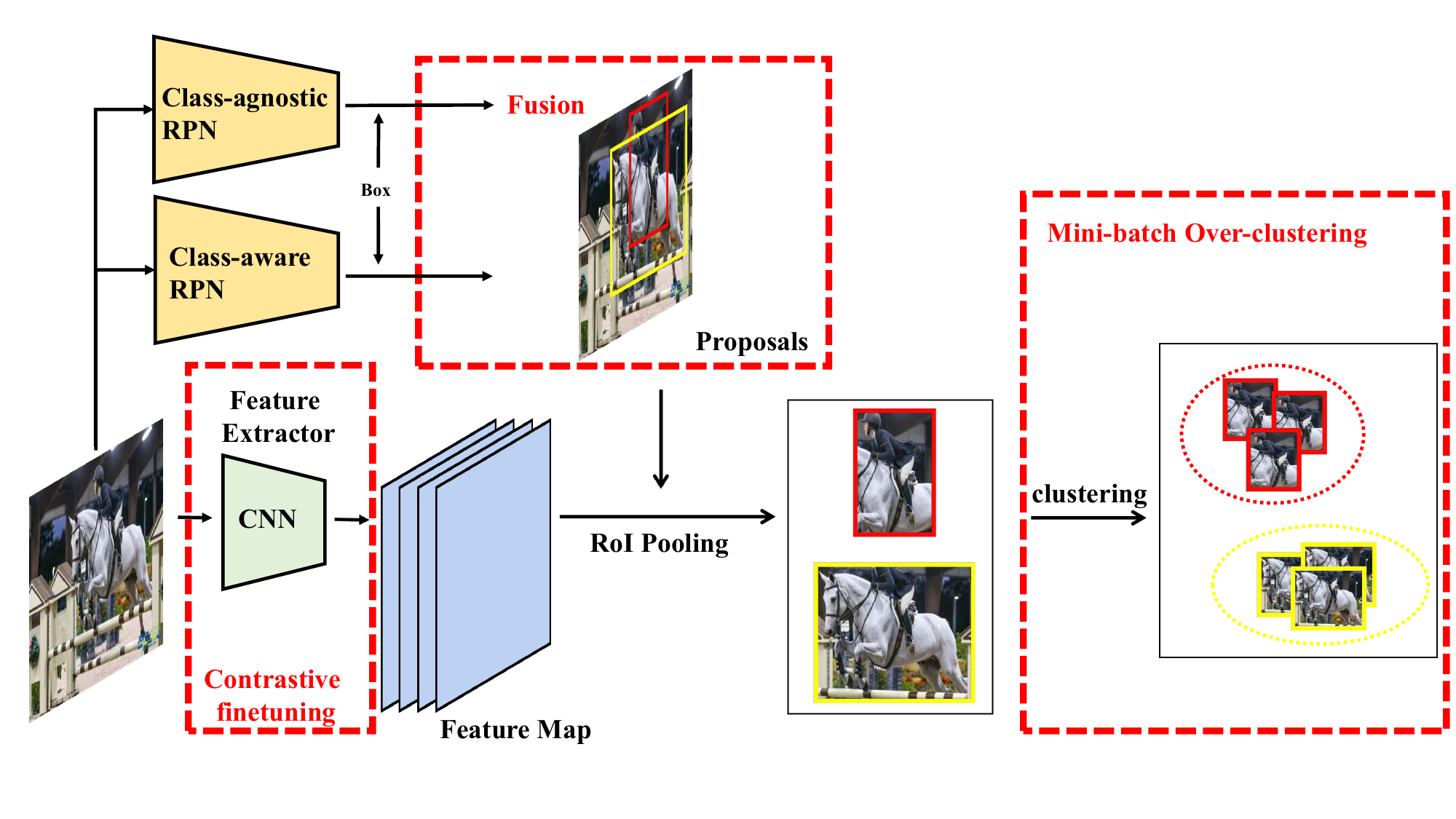}
    \caption{Our Model Architecture. The category discovery and localization pipeline includes three necessary components: 1) learning a feature extractor, 2) discovering boxes from unlabelled images, 3) grouping and categorizing the extracted boxes.}
    \label{fig:Overview}
\end{figure*}

\section{Related Works}
\subsection{Open-set Recognition}
Scheirer~\cite{scheirer2012toward} found that the field of machine vision in real-world scenarios is more of an "open set" recognition phenomenon, in which open set recognition requires identifying known classes while determining to reject unknown classes, and the model gradually rejects unknown classes in the process of recognition. In this paper, the authors discuss the definition of the open set recognition problem as an empirical risk + open space risk minimization problem. Jain~\cite{jain2014multi} further discusses the open set recognition problem in the above work, they believe that the unknown class can be effectively rejected if the known class can be fully modeled, so statistical extreme value theory is used to evaluate the posterior probabilities. Dhamija~\cite{9093355} redefined the open set object detection problem (OSOD), in which the authors argue that the traditional object detection classifier can effectively reduce the model to discriminate unknown classes as known classes with high confidence while proposing a new metric to evaluate the performance of the detector at the time of analysis.

\subsection{Open-world Recognition}
Although the open set recognition can identify unknown categories, the dataset is dynamic for object detection in real scenarios, and the system detects and adds new categories for objects that are continuously inputted. Bendale~\cite{DBLP:journals/corr/BendaleB14} proposed the open world recognition problem and used the nearest non-outlier (NNO) method, by which the model can be effectively optimized and can enable the model to gradually add new categories after balancing the spatial risk of the open world after detecting outliers.
Joseph~\cite{DBLP:journals/corr/abs-2103-02603} extends the theory of the open world to the field of object detection and proposes Open World Object Detection (OWOD) The model achieves incremental learning by a simple and practical method of instant playback, which enables the model to reduce the recognition accuracy of old categories while learning new ones.
Subsequently, for the subsequent work Zhao~\cite{DBLP:journals/corr/abs-2201-00471} improved the experimental setup and thought about more basic principles to guide the construction of the OWOD benchmark. Firstly, the test set for VOC and COCO should contain a perfect unknown label, if there is no complete labeling, the detected images will be considered as False Positive.

However, OWOD and OSOD for the unknown category identification process, the model is known the number of categories of unknown categories to carry out the statistics of unknown categories, when the number of unknown categories in advance will have an impact on the effect of clustering, and in the real open world scenario, we and the model can not know the number of unknown categories in advance, so such a statistical process is still open to question. Moreover, for learning new categories, OWOD and OSOD require additional staff to re-label the new target images, while our model does not need to know the exact number of unknown categories and does not need to re-label the new targets.

\subsection{Category Discovering}
Mining for new classes is also one of the foundations of open-world detection, where new classes are mined for the recognition of localization on unlabeled data in images. In image recognition, the mining of unknown classes starts with region proposals for the search of objects. In the beginning, region search was mainly done by sliding window approach~\cite{6133291} to obtain the boxes of the object, by returning the better boxes as offers or by classifying them by models. However, it has been found that usually in a downstream task for object detection task, the target is usually extracted using low-level features such as texture features, edge features, etc. However, the region extraction network in Faster-RCNN~\cite{DBLP:journals/corr/RenHG015}, which directly uses the features generated in the object detection task, but faster-RCNN will target the prior knowledge of the model to improve the extraction of known categories, is not a good choice for the identification and localization of unknown categories.

\subsection{Novel Category Discovery}
The concept of Novel Category Discovery (NCD) was first introduced in~\cite{hsu2018learning}, where the authors proposed clustering unlabeled samples in a given labeled dataset without class overlap. In~\cite{hsu2018learning}, a prediction network is trained in advance on the labeled data, and then the network predicts pairwise similarity between unlabeled data to discriminate unknown classes and identify new categories. Many subsequent NCD papers~\cite{zhao2021novel,zhong2021neighborhood} have followed a two-step training strategy. In~\cite{zhao2021novel}, a dual-branch learning framework is used to learn from the target data, with one branch focusing on local fine-grained information and the other branch focusing on global image features. Pseudo-labels are generated through dual-ranking on the two branches for training on unknown data. Currently, NCD problems assume that the objects of interest to the model are known (pre-cropped) and do not exist in the training images. However, we believe that in real-world scenarios, the objects of interest to the model should be detected and localized in the images. Therefore, we adopt the Dual RPN approach to extract information about unknown classes from the images and perform localization and recognition, which improves the model's generalization.

\begin{figure*}[t]
    \centering
    \includegraphics[width=0.8\linewidth]{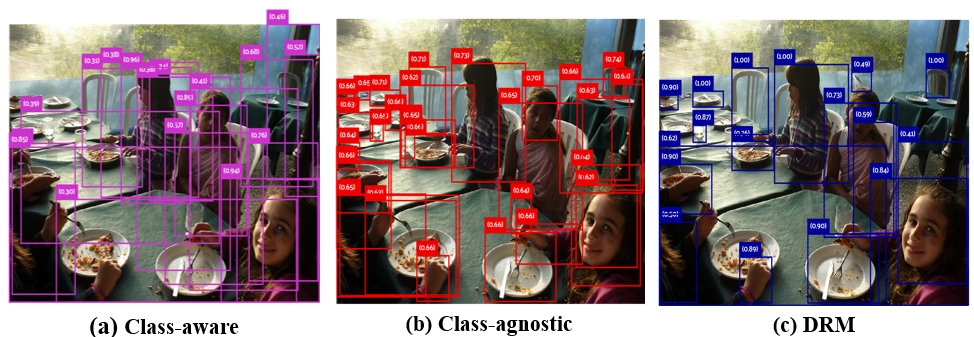}
    \caption{The effect of three kinds of RPN. The object detector in (a) is used a class-aware RPN. The object detector. In (b) uses an object detector that removes the classification head and learns objectness alone. In (c) is our proposed object detector based on IOU regression.}
   \label{fig:RPN}
\end{figure*}

\section{Framework Details}
\subsection{Overview}
In this section, we illustrate our approach in Figure \ref{fig:Overview}. The image passes through two branches: the dual RPN module, which consists of two object detectors that generate boxes for different requirements. Both sets of boxes are processed and used as the final proposal input.
Simultaneously, the image features are extracted by the feature extractor and compared with the extracted image features through semi-supervised contrastive learning. This process helps determine whether the features belong to known or unknown categories. The Region Proposal Network (RPN) extracts image proposals, and the features are then pooled using ROI pooling to associate the proposals with the features.
Finally, the proposals and features are sent to the clustering module, where all instances are clustered. This clustering ensures that instances with similar features are grouped together, allowing for the discovery of different unknown categories.

\subsection{Debiased Region Mining}
In the field of object detection, the Region Proposal Network (RPN) is commonly used for detecting objects in images. The requirement for an object detector is to predict the positions of all the objects of interest in the image. As a two-stage method, the RPN was designed from the beginning of the algorithm with the expectation that it would propose all categories of objects in the image. However, in practical tasks, we have observed two scenarios with RPN. Firstly, when encountering unannotated images, the model tends to classify them as background without localizing any objects. Secondly, when the model recognizes unknown objects, it mistakenly classifies them as known objects with high confidence. In Faster R-CNN, the object localizer is used for the upstream task of the classification head, extracting known classes that the model is interested in, which leads to a preference for recognizing known objects and severely affects the model's generality.

In Figure \ref{fig:RPN}, the utilization of the object extractor in the standard Faster-RCNN for locating image information is evident in (a). This extractor is pretrained on the VOC dataset and subsequently applied to the COCO dataset, with the inferred box serving as the proposal. Such proposals exhibit higher confidence for known objects in the VOC, resulting in improved proposal quality. However, proposals with average confidence tend to be aggregated, often containing only a partial portion of the target object. Consequently, the detected objects demonstrate limited generalization to the categories in the VOC.
In (b), an attempt was made to generate proposals by removing the classification head and solely learning the objectness of the images within the network. Notably, the number of proposals obtained by the network surpasses that of the baseline and necessitates subsequent filtering. Despite an enhancement in proposal generalization compared to the baseline, the accuracy in locating VOC categories remains sub-optimal, and numerous proposals still exhibit clustering.
In (c), we propose a merging approach for the aforementioned boxes. By selecting the box with a reliable confidence region from both boxes and scaling the confidence of each box, the final proposal is unified through non-maximum suppression (NMS). This method significantly improves the quality of the proposals, enabling the extraction of more target objects without compromising the accuracy for known VOC categories. Additionally, it effectively resolves the issue of proposal clustering.

We believe that the NCDL problem in real-world scenarios should be more aligned with the object detection scenario in an open world, where the object extractor should not be limited by classification heads to improve the extraction performance of the RPN. Therefore, we draw inspiration from the ideas proposed by~\cite{DBLP:journals/corr/abs-2108-06753} and others, and introduce a class-agnostic RPN that can generate more generalizable objectness scores and retrieve more objects.
In our approach, we replace the class-related losses with a class-agnostic loss in Faster-RCNN. This loss only estimates the objectness of a region by: 1) using centerness regression instead of the classification loss in the RPN, and 2) using IoU regression instead of the classification loss in the ROI head. By enforcing the detector to focus solely on localization-aware IoU/centerness, the resulting class-agnostic RPN can reduce bias and discover more potential unknown objects.
After obtaining two sets of detection boxes for the same image using the detector, we conducted reliability analysis on these two sets of boxes. The distribution of the boxes on the confidence interval graph is different (detailed in the supplementary material), indicating that each set of boxes has its own advantages and disadvantages. Therefore, we propose a method called "Debiased Region Mining" (DRM) to obtain two different sets of boxes through class-aware RPN and class-agnostic RPN. As shown in the figure, the boxes obtained by the class-aware RPN have higher accuracy on known classes but poor generalization, and they also perform poorly on unknown classes. On the other hand, the boxes obtained by the class-agnostic RPN may not perform as well on known classes as the former, but they have stronger generalization on unknown classes. This inspires us to combine these two sets of boxes to obtain a new set of boxes that combines the advantages of both.

Denote the above two sets of boxes and their confidence scores as $\lambda_1$ and $\lambda_2$. We assume that the scores obey two different distributions $\phi_1 $and $\Phi_2 $, respectively, and we first need to map these two distributions to a uniform $ \Phi $ to remove the gap between different box generation methods. To keep the boxes with high confidence itself and screen out the boxes with very low confidence, we also set the threshold $\alpha_i$, $\beta_i (i=1, 2)$ to filter the confidence levels.
After this, we merged the two sets of boxes and used Class-agnostic NMS to merge the redundant boxes to get the post-fusion results.
\subsection{Semi-supervised Contrastive Finetuning}
After we get the bbox on the Discovery set, we adopt a Semi-Supervised Instance-level Contrastive Learning approach in order to extract more generalized and expressive features.

First, we crop images in VOC dataset into patches according to ground truth boxes to form our labeled set $B_{\mathcal L}$. Then, proposals we generated on COCO train split are used to process patches in unlabeled set $B_{\mathcal U}$.
After that, given two different views of the same patch $\mathbf{x}, \mathbf{x}'$ generated by random augmentations, we formalize the unsupervised contrastive loss as: 
\begin{equation}
    \label{eq:unsupervised-contrastive-loss}
    \mathcal{L}_i^u=-\log 
    \frac{\exp(\mathbf{z}_i\cdot\mathbf{z}_i'/\tau)}{\sum_n\mathit{1}_{[n\not=i]}\exp(\mathbf{z}_i\cdot\mathbf{z}_n/\tau)}
\end{equation}
where $\mathbf{z}, \mathbf{z}'$ are corresponding feature and $\tau$ is temperature hyperparameter.

For images in supervised dataset, annotations are utilized to form supervised contrastive loss:
\begin{equation}
    \mathcal{L}_i^s=-\frac{1}{|\mathcal{N}(i)|}\sum_{q\in \mathcal{N}(i)}\log \frac{\exp(\mathbf{z}_i\cdot\mathbf {z}_q/\tau)}{\sum_n\mathit{1}_{[n\not=i]}\exp(\mathbf{z}_i\cdot\mathbf{z}_n/\tau)},
    \label{eq:supervised-contrastive-loss}
\end{equation}
Where $\mathcal{N}(i)$ denotes the indices of which has the same label as $\mathbf{x}_i$. Finally, the total loss is constructed as following:
\begin{equation}
    \mathcal{L}^t=(1-\lambda)\sum_{i\in B}\mathcal{L}_i^u+\lambda\sum_{i\in B_{\mathcal L}}\mathcal{L}_i^s.
    \label{eq:total-contrastive-loss}
\end{equation}
This loss will supervise the training of our feature extractor.

\begin{table*}\small%
\centering
    \begin{tabular}{c|cccc|cccc}
\hline
\multicolumn{1}{l|}{Method}    & AuC@0.5 & AuC@0.2 & \#Disc. Cls. & CorLoc & AuC@0.5& AuC@0.2 & \#Disc. Cls. & CorLoc                           \\ \hline
\#cluster & \multicolumn{4}{c|}{80} & \multicolumn{4}{c}{1000} \\ \hline
Vplow & 0.0548  & \textbf{0.1229}  & 18 & 0.9413 & 0.0818 & 0.0961 & 40 & 0.9413 \\
DualMem & 0.0503  & 0.1174  & 16 & 0.9433 & 0.0025 & 0.0055 & 42 & 0.0746 \\
OLN & 0.0517  & 0.1076 & 18 & 0.9713  & 0.0971  & 0.1427  & 45 & 0.9713 \\ 
\textbf{Ours} & \textbf{0.0613} & 0.1203 & \textbf{20} & \textbf{0.9934}& \textbf{0.1052} & \textbf{0.2381} & \textbf{57} & \textbf{0.9934} \\ \hline
\end{tabular}
\caption{Comparison with contemporary discovery methods using AuC and CorLoc for unknown classes in COCO train2014. }
  \label{tab:overview}
\end{table*}

\subsection{Clustering}
After completing the contrastive learning for unknown category objects, the model performs clustering analysis on the obtained information, aggregating unknown images with similar feature representations into clusters. We initially use a clustering method similar to K-means but with two modifications. Firstly, we employ an over-clustering strategy by forcing the generation of another finer-grained partition of unlabeled data and increasing the number of K (estimated number of clusters) to improve clustering purity and feature representation quality. Over-clustering is beneficial in reducing supervision, allowing the neural network to decide how to partition the data. This separation is effective in the presence of noisy data or when intermediate classes are randomly assigned to neighboring categories.
Secondly, using K-means in our task is time-consuming. So we use Minibatch K-means, which is an optimization algorithm for K-means in large-scale data, to reduce training computation time by randomly sampling a subset of data during training while optimizing the objective function.

The main steps of the algorithm are as follows: 1) Initially, a subset of training data is extracted, and K cluster centers are constructed using K-means. 2) Sample data is further extracted from the training set and added to the model, assigning them to the nearest cluster center. 3) The cluster centers of each cluster are updated. 4) Iteratively repeat steps 2 and 3 until the cluster centers stabilize or reach the maximum number of iterations.

\begin{figure*}[t]
  \centering
    \includegraphics[width=0.8\linewidth]{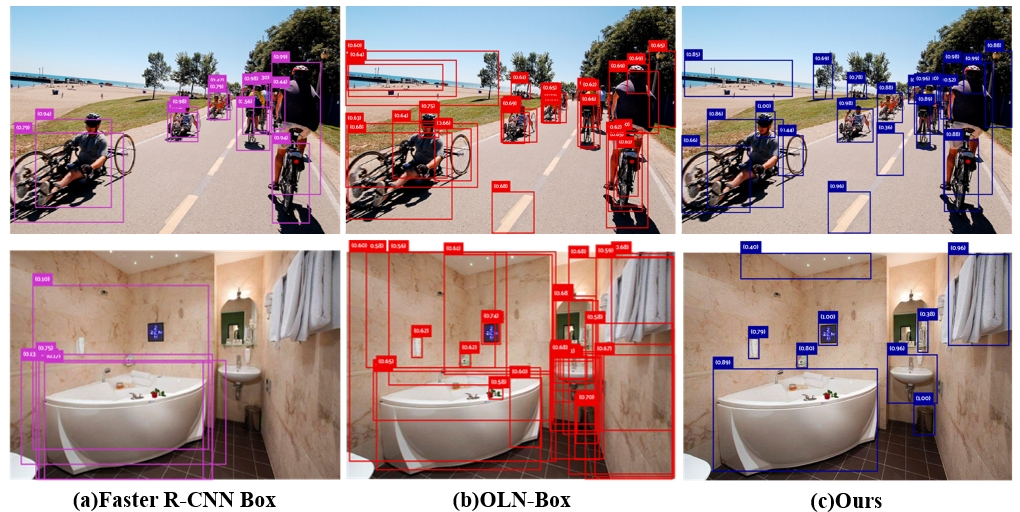}
   \caption{We validate our method on the COCO dataset and present the results of its visualization (in the visualization results we remove the boxes with too low confidence). It can be clearly seen that our method can effectively localize more target objects than the standard Faster R-CNN. Compared with OLN-Box, our method is more accurate in locating known classes.}
   \label{fig:Fusion}
\end{figure*}

\section{Experiments}
\subsection{Datasets}
The datasets we mainly used are (1) the Pascal VOC 2007 dataset~\cite{everingham2010pascal} contains 10k images with 20 labeled categories. The dataset is mainly focused on classification and detection tasks for the targets. (2) The COCO 2014 dataset~\cite{lin2014microsoft} contains 80 annotated categories with 80k images in the training set and 5k images in the validation set.

In our experiments, we consider the 20 categories in the VOC dataset as known categories for our model, while the remaining 60 categories in the COCO dataset, which do not overlap with VOC, are treated as unknown categories to represent unfamiliar images in the open world. We train our dual RPN using the VOC dataset to provide prior knowledge to the target extractor. However, the model cannot learn the remaining 60 unknown categories in advance.

Extending the model to handle these 60 unknown categories presents a challenging problem. Firstly, the images in the COCO dataset exhibit greater diversity and complexity compared to those in the VOC dataset. Each image in COCO contains multiple categories of targets and a complex environment. Secondly, replicating a real-world target detection setup is not feasible. Although the VOC and COCO datasets have similar category distributions, the COCO dataset visually encompasses all the labels of the 20 categories in VOC. This allows for a comprehensive evaluation of the model's ability to discover new categories.

\begin{figure*}[t]
  \centering
    \includegraphics[width=0.7\linewidth]{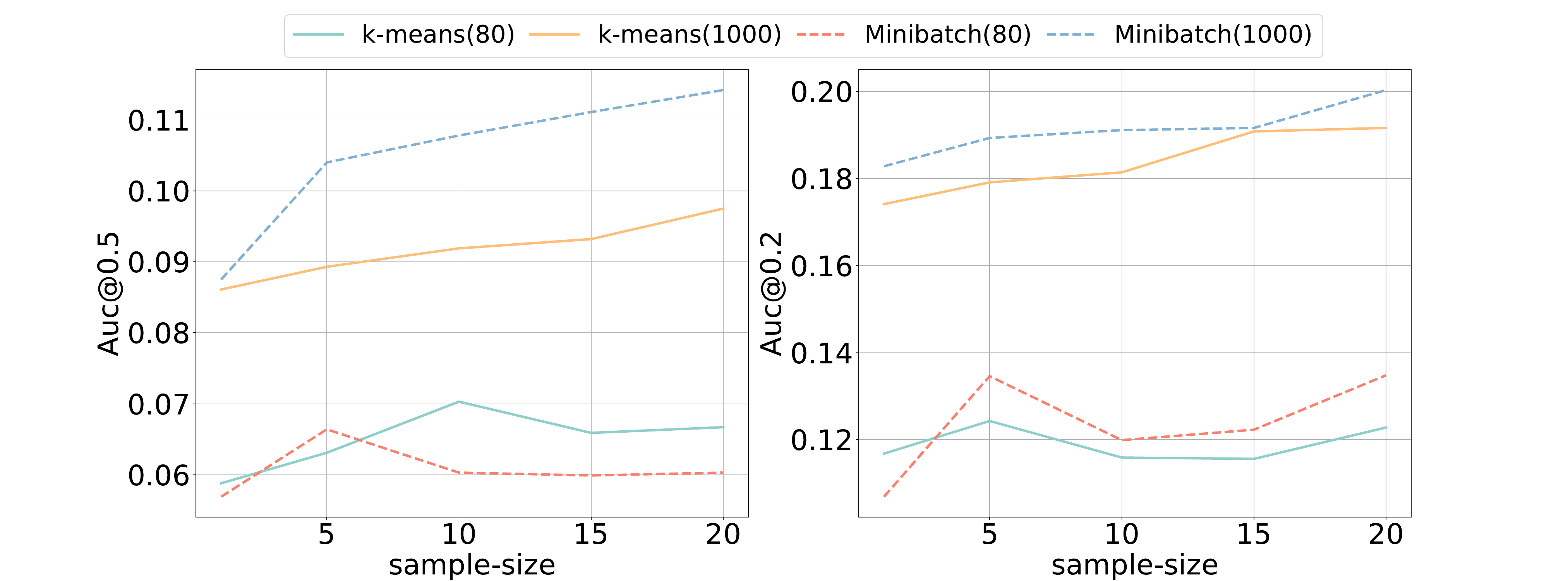}
   \caption{The impact of the two clustering methods on AuC with varying sample sizes and total number of clusters.}
   \label{fig:AuC}
\end{figure*}

\subsection{Experimental Metrics}
We summarize the metrics used in our experiments.

\textbf{Localization Metrics}: We employ the CorLoc metric to evaluate the accuracy of object localization. CorLoc (correct localization)~\cite{DBLP:journals/corr/ChoKSP15} measures the percentage of images where the intersection over union (IoU) between the predicted bounding box and the ground truth bounding box is greater than 0.5. The CorLoc metric assesses the ability of the RPN to accurately localize objects but does not reflect the model's capability to discover new categories.

\textbf{Discovery Metrics}: Mean Average Precision (mAP) is a commonly used metric in object detection to measure the accuracy of object detectors such as Faster R-CNN and SSD~\cite{DBLP:journals/corr/LiuAESR15}. It calculates the average precision value across different recall levels from 0 to 1. By computing mAP, we can evaluate whether the model successfully detects the target objects. However, for the clustering of unknown categories, we need a metric to assess the clustering effect. Therefore, we utilize the area under the curve (AuC) mentioned in~\cite{doersch2014context} to indicate the purity and coverage of the discovered objects. In~\cite{doersch2014context}, it was observed that clusters with a coverage value below 0.5 are usually not target objects. Thus, a higher AuC value indicates higher purity of the target objects in the corresponding clusters, suggesting that the objects within the target clusters are more likely to belong to the same class. Additionally, we report the number of unknown classes discovered by the model to evaluate its effectiveness in mining new classes in an open world. Together, these metrics provide a comprehensive evaluation of the model's performance in an open world setting.

\subsection{Results and Analysis}
We compared our method with DualMem~\cite{DBLP:journals/corr/abs-2105-01652}, Vplow\footnote{https://rssaketh.github.io/project
pages/obj disc new.html}, OLN~\cite{DBLP:journals/corr/abs-2108-06753}. Vplow replaces the clustering method with K-means, while the remaining steps are the same as DualMem. We also compared clustering with 80 categories and clustering with 1000 categories to demonstrate the superiority of the over-clustering strategy. Since the K-means algorithm does not scale well to large datasets like COCO, we randomly sampled 10,000 instances from all input instances and fed them into the model, then made predictions on all images to obtain the final clustering results. In Table \ref{fig:Overview}, we report the AuC, the number of discovered unknown categories, and the CorLoc localization metric on the COCO 2014 training set to evaluate the overall performance of different methods in the large-scale setting. Since DualMem does not provide open-source code, we reproduced its dual-memory mechanism based on the description in the paper. From Table \ref{tab:overview}, we can see that our method outperforms the baseline methods in all metrics under the same settings.

For the \textbf{localization metric}, Vplow and DualMem both use Faster R-CNN to extract object proposals from images, while OLN uses a class-agnostic detector for object region extraction. From the data and Figure \ref{fig:Fusion}, we can observe that all methods perform well in object localization. However, Vplow and DualMem have biases in box extraction, performing much better in localizing known categories than unknown categories, and there may be omissions and misjudgments in localizing unknown categories. To address this issue, DualMem introduces low-confidence proposals into the network's input to improve localization and clustering when the number of unknown categories increases. Although this approach allows DualMem to discover more categories, other metrics are significantly reduced (as DualMem does not have a specific cluster number setting, we set the memory slots based on the clustering categories). Overall, our proposed debiased classification method not only effectively explores unknown categories but also improves the localization of known categories.
For the \textbf{AuC metric}, Vplow and DualMem use a DINO-initialized ViT as the feature extractor for instances. Due to the severe clustering bias in the instance localization of previous Vplow, the extracted proposals naturally have similar features, resulting in purer clustering in K-means. Therefore, compared to DualMem, Vplow performs better in terms of AuC.

\begin{table}[ht]\small%
  \centering
  \begin{tabular}{c|ccc}
    \hline
    Method (\#cluster=1000) & AP & AP50  \\
    \hline
    Faster R-CNN box + DINO feature & 1.6 & 4.8\\
    Faster R-CNN + Contrastive feature & 1.9 & 5.2 \\
    OLN-Box  + DINO feature & 1.5 & 4.0 \\
    OLN-Box  + Contrastive feature & 1.5 & 4.2 \\
    DRM box + DINO feature & 3.4 & 7.7 \\
    \textbf{DRM box + Contrastive feature} & \textbf{8.7} & \textbf{20.19}  \\
    \hline
  \end{tabular}
  \caption{Compare the improvement in mAP values for novel category recognition in object discovery on COCO test2014 for each module. }
  \label{tab:map}
\end{table}

\subsection{Ablation Study}
\textbf{Analysis for mAP}
Table \ref{tab:map} presents the impact of using different methods at different stages on the experimental results, with Minibatch K-means being the unified clustering method. From Table \ref{tab:map}, it can be seen that the effects of using different detectors' boxes as input proposals for object localization are consistent with our expectations. The use of fused boxes as input proposals significantly improves target localization. Additionally, the baseline performs better for known categories in VOC, while replacing the classification head yields poorer results for known categories in VOC but can enhance the generalization ability for other unknown objects. For known categories in VOC, the baseline performs poorly but shows improved generality for other unknown objects. Using features extracted from contrastive learning is also better for the model compared to using features extracted solely from DINO, and the experiments demonstrate that our proposed module significantly enhances the performance of the model.

\textbf{Analysis for time of cluster}
Table \ref{tab:time} demonstrates the clustering time performance of different clustering methods under varying numbers of clusters and total sample sizes. The table reveals that for the K-means clustering method, the clustering time significantly increases as the sample size grows, and increasing the number of clusters also leads to a noticeable increase in clustering time. On the other hand, for the Minibatch K-means clustering method, it can significantly reduce the time cost of clustering while having a minor impact on the clustering time with an increase in the number of clusters and sample size.

Figure \ref{fig:AuC} shows the accuracy results of clustering methods with varying numbers of clusters and sample sizes. The over-clustering method improves cluster purity for similar clustering with the same sample size. Increasing the sample size slightly enhances the AuC index with the same number of clusters. Additionally, the Minibatch K-means method dynamically updates cluster centers based on input instances, saving time compared to regular K-means and better approximating complex real-world scenarios.

\begin{table}\small%
  \centering
  \begin{tabular}{l|cc}
    \hline
    Method (\#cluster) & Sample Size & Time (s)  \\
    \hline
    K-means (80) & 10k & 16.371  \\
    K-means (80) & 50k & 127.162 \\
    K-means (80) & 100k & 209.706 \\
    K-means (1000) & 10k & 141.129 \\
    K-means (1000) & 50k & 675.414 \\
    K-means (1000) & 100k & 1864.456 \\
    Minibatch (80) & 10k & 3.316 \\
    Minibatch (80) & 50k & 3.661 \\
    Minibatch (80) & 100k & 3.997 \\
    Minibatch (1000) & 10k & 17.282\\
    Minibatch (1000) & 50k & 23.436\\
    Minibatch (1000) & 100k & 24.026\\
    \hline
  \end{tabular}
  \caption{Effect of different sample size and total number of clusters on clustering time.}
  \label{tab:time}
\end{table}
\section{Conclusion}
In this paper, we propose a new region proposal named Debiased Region Mining (DRM) and a Semi-Supervised Instance-level Contrastive Learning approach. In the former, we use a dual RPN to extract the target objects in the image and use DRM to process them, obtaining a box with accurate localization of the known classes and good generalization to the unknown classes as the proposal. In the latter, we employ semi-supervised Instance-based Contrastive Learning to extract image features with better generalization and expressiveness. We have conducted extensive experiments to validate our claims. The experiments demonstrate that our approach currently outperforms other novel class mining and localization algorithms in the open world. Our work provides an idea for target detection in realistic and complex environments.

\section{Acknowledgements}
This work was supported by National Key R\&D Program of China (No. 2022ZD0119401), National Natural Science Foundation of China (No. 61925107, U1936202, 62021002, 62301066), Key R\&D Program of Xinjiang, China (No. 2022B01006), Zhejiang Provincial Natural Science Foundation of China under Grant (No. LDT23F01013F01) and Tsinghua-OPPO JCFDT.

\end{document}